%% file: paper_acm.tex
\documentclass[sigconf, nonacm, screen]{acmart} %

\usepackage{hyperref}

\usepackage{algorithm}%
\usepackage{algpseudocode}%

\usepackage{multicol}   %

\begin{document}

\title[Knowledge Extraction Using Language Model Chains]{Fast and Accurate Contextual Knowledge Extraction Using Cascading Language Model Chains and Candidate Answers}

\author{Lee Harris}
\orcid{0009-0008-9260-5029} %
\affiliation{%
  \institution{The University of Kent}
  \city{Canterbury}
  \state{Kent}
  \country{UK}
}
\affiliation{%
  \institution{TMLEP \& Elysium Web Services}
  \city{Ashford}
  \state{Kent}
  \country{UK}
}

\begin{abstract}
	\input{sections/abstract}

\end{abstract}



\graphicspath{{figures/}} %

\maketitle

\section{Introduction}
\input{sections/introduction}

\section{Language Model Chains}
\input{sections/algorithm}

\section{Methodology}
\input{sections/methodology}

\section{Experiment 1: language models}
\input{sections/experiment_1}

\section{Experiment 2: language model chains}
\input{sections/experiment_2}

\section{Discussion}
\input{sections/discussion}

\section{Conclusion}
\input{sections/conclusion}

\begin{acks}
  \input{sections/acknowledgements}

\end{acks}

\bibliographystyle{ACM-Reference-Format}
\bibliography{bib}

\end{document}

%% file: sections/abstract.tex
Language models can capture complex relationships in given text, but these are notorious for being costly and for producing information that does not exist (i.e., hallucinations).
Furthermore, the resources invested into producing this information would be wasted if it were incorrect.  %
We address these issues by proposing, implementing, and applying the Language Model Chain (LMC) algorithm.
In this, a language model's response to a given prompt about given text is only correct if it exists in the collection of possible (i.e., candidate) answers, and text corresponding to incorrect responses is fed into a more predictive (but slower) language model. 
This process is repeated for a collection of language models, or until all predictions about the text are correct.
We used the LMC algorithm to extract patient dates of birth from medical documents, and combining a collection of language models in a multi-stage cascade significantly increased prediction speed and accuracy over individual language models, while greatly reducing the number of corresponding hallucinations.
We believe that the novel LMC algorithm significantly contributes to the knowledge extraction field, and that this should be explored much further in the future.

%% file: sections/keywords.tex
Healthcare, Language Models, Multi-stage Cascade Classification, Language Model Chains, Information Retrieval, Generative AI, AI Computational Optimisation

%% file: sections/introduction.tex
Generative deep learning is increasingly being used to extract knowledge from data \cite{ye2022generative}, and Language Models (LMs) which are able to extract complex and contextual patterns from text are amongst the most popular instances of these \cite{chang2024survey}.
However, processing large amounts of text (especially without GPU-accelerated hardware) is often slow \cite{buber2018performance}, and it is well documented that these models tend to fabricate answers when they do not know how to respond (i.e., hallucinations) \cite{farquhar2024detecting}.
Additionally, generative AI is an energy-intensive activity \cite{luccioni2024power} that is often considered to be extremely harmful to the environment \cite{wu2022sustainable}, and natural language processing has not experienced the same performance boost that the You Only Look Once (YOLO) algorithm brought to computer vision \cite{redmon2016you,lee2024comparative}.

Our task was to extract Dates Of Births (DOBs) from a corpus of medical documents.
These contained many dates, and it was rarely possible to extract a patient's DOB without understanding the context that each potential date was mentioned in. 
For instance, a few of the dates that a medical document may contain are a patient's DOB, the DOBs of their siblings, appointment dates, procedure dates, diagnosis dates, medicine manufacture and expiration dates, correspondence dates, record creation and modification dates, and text that looks like dates. %
We were also required to process the medical documents on a local device that did not contain a GPU. 
We tried to solve this problem through the use of regular expressions \cite{harris2025extracting}, fine-tuning foundation models \cite{vm2024fine}, and Named Entity Recognition (NER) \cite{li2020survey} (these explorations are not described in the current research paper), but these did not produce the desired accuracy, and performing deep learning inference within the constraints of our hardware was very challenging.

We were best able to solve the task by proposing, implementing, and applying the novel Language Model Chain (LMC) algorithm.
This combines sequential application of \textit{`good enough'} LMs with validation of their responses against possible (i.e., candidate) answers to very quickly extract contextual knowledge from a corpus of text that contains multiple competing answers.
We explored twelve popular and small open-source LMs due to the resource and sensitive-data constraints, and a subset of these were used to produce four LMCs. 
These were shown to be significantly faster and perform noticeably better than the LMs that they contained. 
We also found evidence that a negative correlation exists between the speed of an LM and its predictive performance, and that LMs containing many parameters do not necessarily perform better than LMs containing fewer. %

LM research is a very active field.
\cite{rajeev2025cats} recently explored the delegation of simpler tasks to smaller LMs (opposed to difficult tasks and large LMs), although they focused on the problem of adversarial attack in computer vision, and their method was not applied to real-world data. %
\cite{shojaee2025illusion} stated that LMs should be used to solve precise and well-defined tasks as these are unable to reliably handle complex data, while \cite{belcak2025small} argued that smaller LMs are better suited to solving well-defined and precise tasks that need to be executed repeatedly. %
Other researchers have explored ways to constrain LM output using regular expressions and distinguishable patterns \cite{kuchnik2023validating}.
A large contributor to these works, and to the LMC algorithm, are the recent increases to LM context size as these have enabled larger and more complex Retrieval Augmented Generation (RAG) applications \cite{fan2024survey}.

We believe that the LMC algorithm contributes practically and significantly to the field of knowledge extraction. 
However, these experiments were not an exhaustive exploration of LMs, LMCs, prompt engineering (or its encompassing topic; context engineering \cite{zhang2025knowledge}), or candidate answer extraction, and these should be explored further in the future.

%% file: sections/algorithm.tex
\begin{figure}
  \centering
  \includegraphics[trim={-2cm 0 -2cm 0},clip,width=1.0\linewidth]{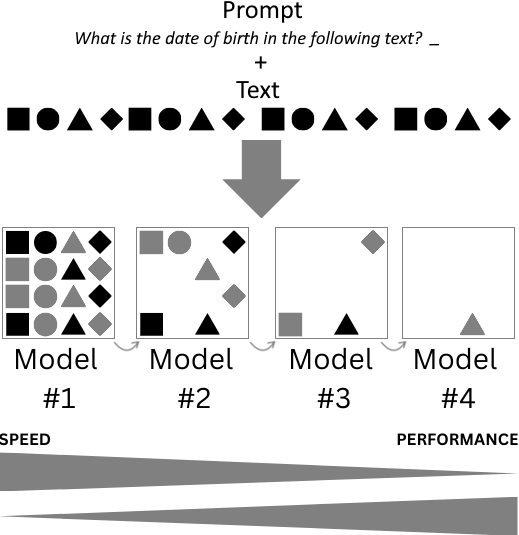}
  
  \caption{
 	A high-level visualisation of how the Language Model Chain (LMC) algorithm would apply to a given prompt and given text. These hyperparameters are used for illustrative purposes, and can be changed. An incomplete RAG prompt that states ``What is the date of birth in the following text? \_'' is iteratively completed by 16 text examples (which are represented by the squares, circles, triangles and diamonds), and these are fed to the first (language) model. The grey boxes containing the black and grey shapes show the examples that are fed into each model, and the small grey arrows indicate the flow of the sequence (i.e., from left to right). The grey shapes indicate examples whose model response was correct, and the black shapes indicate those that were sent onto the next model. The first model correctly predicted 9 data examples, and the 7 data examples whose model responses did not match a pattern of text in the respective data example were fed to a more predictive (but slower) language model. This process was repeated until all data examples were correct. The second model correctly predicted 4 data examples, the third model correctly predicted 2 data examples, and the fourth model correctly predicted 1 data example. The triangles at the bottom of the figure show the hypothesised negative correlation that exists between computational speed and predictive performance. 
    }
  
    \Description[A high-level visualisation of the LMC algorithm]{Figure fully described in the caption.}
  
  \label{FIG:algorithm_image}
\end{figure}

The Language Model Chain (LMC) algorithm is conceptually simple.
First, all potential (i.e., candidate) answers to a target query are extracted from given text.
We currently use regular expressions for this. %
Next, the first item in a stack of LMs is popped off and prompted about the text.
Finally, this process is repeated for any text whose corresponding LM response does not exist in the collection of candidate answers, or until all of these are correct.
Alg.~\ref{ALGORITHM:algorithm_pseduocode} presents LMC pseudocode, and Fig.~\ref{FIG:algorithm_image} presents a visualised example of this algorithm.

\begin{algorithm}
  \caption{The Language Model Chain (LMC) algorithm: validating successive Language Models (LMs) predictions using a set of candidate answers %
to quickly and accurately extract contextual knowledge from a collection text.}
    
	\textbf{Input x: data to predict} \\
	\textbf{Input prompt: a language model prompt} \\
	\textbf{Input lms: a stack of language models} \\	
	\textbf{Output: predictions of each data example} 
  \begin{algorithmic}[1]
    \Procedure{LMC}{$x$, $prompt$, $lms$}
    	
      	\State $model \gets lms.pop()$ \Comment{pop() returns and removes the first item of a stack}
      	\State $y \gets predict(x, prompt, model)$
      	\State $y^c \gets get\_candidate\_answers(x)$ %
      	\State $x^i \gets get\_incorrectly\_predicted\_data(x, y, y^c)$
      	\State $y^{lmc} \gets LMC(x^i, prompt, lms$)
      	\State $y \gets merge(y, y^{lmc})$
      \\
      \Return $y$ %
    \EndProcedure
  \end{algorithmic}
  
    \label{ALGORITHM:algorithm_pseduocode}
    
\end{algorithm}

Incorrect predictions during real-world deployment are typically seen as sunk costs \cite{brynjolfsson2022turing}, but it is irrelevant whether a prediction is slightly or massively wrong: the outcome is the same.
The LMC algorithm uses this information in future predictions, and this is a particular instance of the multi-stage cascade framework \cite{alpaydin1998cascading,viola2001rapid}. 
It is not a variant of the Mixture-of-Experts (MoE) \cite{shazeer2017outrageously} algorithm or the federated learning  \cite{zhao2018federated} paradigm.
Cascade algorithms are often used in computer vision to reduce the incredibly large input space \cite{openWorldRecognition2019,cai2019cascade}, and combining  the use of language models and candidate answers reduces the size of the input space in text problems.

It is possible to imagine the LMC algorithm as a particular instance of a larger `\textit{Model Chain}' framework.
The regular expressions could be replaced by an algorithm that is able to capture more open-ended candidate answers (e.g., NER), the text could be replaced by a different type of data (e.g., pixels or timeseries), and the use of LMs could be replaced by an algorithm that is able to capture broad patterns or accomplish ambiguous tasks (e.g., AI agents \cite{xi2025rise}).
The deterministic structure of regular expressions and the ability to capture candidate answers is a limitation of the LMC algorithm, but these possible replacements greatly increase the practical application and extension potential of this in the future.

%% file: sections/methodology.tex
\subsection{Data Description}
We explored 2327 medical documents that were collected over a decade, and these consisted of various personal details (names, addresses, etc.) and medical histories (ailments, prescriptions, medications, appointments, notes, etc.). %
The text in these was digitally written in popular (e.g., Times New Roman, Arial, Calibri, etc.) typefaces from left to right on vertically or horizontally oriented white backgrounds, and these were stored in either portable (i.e., .pdf) or Microsoft Word (i.e., .docx) format. 
The text was extracted from documents with a .docx extension using the mammoth (\url{https://pypi.org/project/mammoth/}) python package, and it was extracted from  the text layer of each PDF using the pypdfium2 (\url{https://pypi.org/project/pypdfium2/}) python package.
The high-performing Tesseract \cite{smith2007overview} (\url{https://github.com/tesseract-ocr/tesseract}) Optical Character Recognition (OCR) tool was used to extract a digital text transcription in the rare instance that a PDF's textual layer was empty or corrupt. 
Horizontally-oriented pages were rotated to a vertical position.
Leading and trailing whitespaces and empty lines were removed.
Tab.~\ref{TAB:descriptive_data_statistics} shows descriptive statistics about the medical documents.

\begin{table}[h]

		\caption{Descriptive statistics about the 2327 medical documents that were explored. The number before a $\pm$ symbol indicates a document's mean, and the number after it indicates the respective standard deviation. The estimated text comprehensibility was produced by the textstat (\url{https://github.com/textstat/textstat}) tool. The dates were extracted using regular expressions, and the target DOBs were assigned by company experts.}

	\begin{tabular}{lr}
	\hline
	count & 2327 \\
	pages & 29.07 $\pm$ 9.25 \\
	words & 12058.71 $\pm$ 9193.34\\
	estimated suitable reading age & 17-18 (12$^{th}$-13$^{th}$ US grade)\\
	number of day/month/year dates & 3.73 $\pm$ 1.25 \\ 
	\hline
	\end{tabular}

	\label{TAB:descriptive_data_statistics}
\end{table}

\subsection{Language Models}
This research used the open source ollama (\url{https://ollama.com}) \cite{grattafiori2024llama} tool to access a repository of open source LMs. 
These were:

\begin{multicols}{2}
\begin{itemize}
	\item{deepseek-r1:1.5b}
	\item{deepseek-r1:7b}
	\item{deepseek-r1:8b}
	\item{gemma3:1b}
	\item{gemma3:4b}
	\item{gemma3:12b}
	\item{llama3.2:1b}
	\item{llama3.2:3b}
	\item{phi4:14b}
	\item{qwen3:0.6b}
	\item{qwen3:1.7b}
	\item{qwen3:4b}
\end{itemize}
\end{multicols}

\noindent{}%
where $n$ in ':$n$b' indicates the number of trainable parameters in the respective LM.
In many cases, the LMs that were explored were the smallest version of a benchmarked LM so that these were able to execute on the available hardware, although the decision of which LMs to use were ultimately decided by this research paper's authors. 
The `\textit{temperature}', \textit{`random\_seed'}, and `\textit{repeat\_last\_n}' hyperparameters were set to 0 for all LMs. 
All data was stored and processed on a local device, and this was an Apple Macbook Pro 2023 that operated on MacOS v15.5 (Sequoia), and contained an M2 Apple-Silicon processor and 32gb of RAM.
The device was often in daily use alongside obtaining the LM predictions.

\subsection{Date Recognition}
\label{SSEC:date_recognition}
A sequence of characters were recognised as a potential date if they could be converted to day/month/year (DD/MM/YYYY) format, and this was accomplished by using bespoke regular expressions.
Some blocks of text explicitly contained a number between 1 and 31 (inclusive), followed by a separating character, followed by a number between 1 and 12 (inclusive), followed by a separating character, which was finally followed by a 2 or 4 digit number. 
A 2-digit number greater than 25 was identified as a year in the 20$^{th}$ century, otherwise it was identified as a year in the 21$^{st}$ century. 
The two separating characters needed to be the same character from the set \{/.-\}.
As is common, some dates were written in similar formats to `the \_ of \_, \_' (e.g., `\textit{the 5$^{th}$ of May, 1998}', `\textit{11$^{th}$ of Jun 62}', etc.), and these were successfully captured and converted by replacing the individual tokens in possible date patterns with their date counterparts (e.g., `\textit{of Jun}' became `/06/', `$11^{th}$' became `11', and `\textit{62}' became `1962'). 
Finally, all blocks of text that were provisionally identified as dates 
were transformed into a python datetime (\url{https://docs.python.org/3/library/datetime}) object, and invalid objects were ignored.

\subsection{Predictive Performance}
The ability of the LMs and LMCs to predict dates and DOBs are reported using the precision, recall, and F$_1$ (score) statistics. 
If target DOBs are present, then precision indicates how likely an LM or LMC is to predict these, while recall indicates how many of these predictions are correct. %
If not, then precision indicates how likely an LM or LMC is to predict a date, and recall indicates whether these actually exist in the text (i.e., hallucination rate).
The F$_1$ score represents the harmonic mean (i.e., the average predictive performance) between these, and higher values are better than lower values.
Formally, these are 

\begin{equation}
	precision = 100\frac{TP}{TP + FP}, \\
\end{equation}

\begin{equation}
	recall = 100\frac{TP}{TP + FN}, \\
\end{equation}

\begin{equation}
	F_{1}\_score = 2 \times \frac{precision \times recall}{precision + recall},
\end{equation}

\noindent{}%
where $TP$ are the True Positives, $FP$ are the False Positives, and $FN$ are the False Negatives. These terms exist in the set of positive natural numbers ($\mathbb{Z}^+$), and equations 1-3 are percentages between 0 and 100. %

\subsection{Tokens Per Second}
The industry-standard Tokens Per Second (TPS) metric was used to report the computational speed of each LM and LMC.
This divided the total characters in a given document by the size of a token and the amount of seconds that it took to produce a prediction about the document.
OpenAI\footnote{\url{https://help.openai.com/en/articles/4936856-what-are-tokens-and-how-to-count-them}, last visited on: 11$^{th}$ of July, 2025} define a token as 4 characters.  
The formula for this is 

\begin{equation}
	TPS(D, M, P) = \frac{D_c}{4 \times seconds(D,M,P)},
\end{equation}

\noindent{}%
where $D$ is a document, $D_c$ is the number of characters in $D$, $M$ is either an LM or LMC, $P$ is a prompt, and $seconds$ is a function that produces the number of seconds it took for $M$ to create a completed response about $D$ when prompted with $P$.

\subsection{Ethical Considerations and Compliance} 
This applied research explored real-world patient and company data. 
The predictions that were made were for internal company use only, and these were not used to produce automatic decisions or actions.
The target DOBs in each document were manually assigned by approved company experts.
The original documents were not altered by the prediction process.
All data was stored and processed on the local device, and viewed and manipulated by humans and machines on the company premises by those who were authorised to do so.
This project was ethically approved by the author's university, and the patients consented for the company to store their medical records. 
The company strictly and rigidly adheres to the ISO 27001:2022 %
standard, and the secure and expert storage of data.

%% file: sections/experiment_1.tex
The first experiment benchmarked the predictive performance of the LMs. %
We began by extracting all of the candidate dates from the medical documents using regular expressions (see Sec.~\ref{SSEC:date_recognition}), and then we prompted each LM about the respective text.
The prompt was
\\

\noindent{}%
\textbf{Prompt:} \textit{TEXT: \_. QUESTION: What is the patient's date of birth? The date must be in DD/MM/YYYY format. Return 'I do not know' if the date of birth is not written in the TEXT.}
\\
 
\noindent{}%
where each document replaced `\_' by the respective text.  
We then extracted the first date from each response (ignoring any `thinking' stages), and we checked whether this was in the collection of candidate dates.

\begin{figure}
  \centering
  LM Tokens Per Second
  \\
  \includegraphics[trim={0 0 0 0},clip,width=0.9\linewidth]{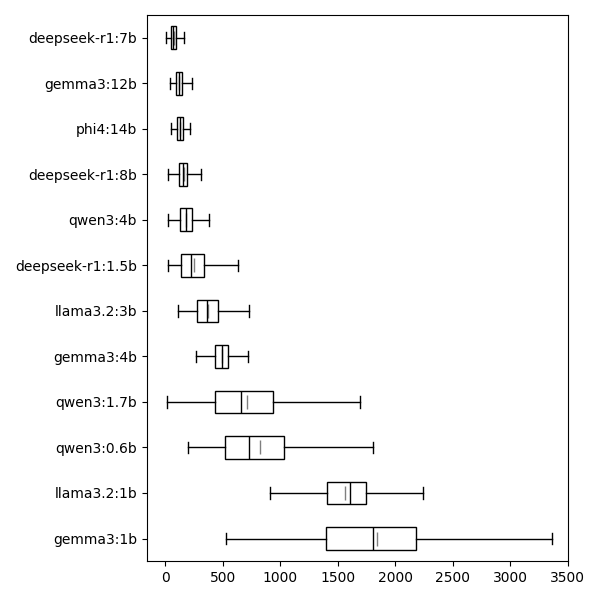}
  
  \caption{
  The tokens per second of each LM when the task was to extract a patient's DOB from the medical documents.  Each box contains six vertical lines.   The first two (left-right) respectively represent the lowest value and the first quartile.  The last two respectively represent the highest value and the third quartile.  The grey line inside each box represents the mean, and the black line represents the median. The LMs are ordered according to their mean tokens per second, with the smallest value at the top, and the largest at the bottom.
  }
    
  \label{FIG:lm_tokens_per_second}
  
    \Description[Each language model's mean tokens per second]{
  The majority of the figure is described in its caption.
  However the y-axis labels (from top to bottom) are: deepseek-r1:7b, gemma3:12b, phi4:14b, deepseek-r1:8b, qwen3:4b, deepseek-r1:1.5b, llama3.2:3b, gemma3:4b, qwen3:1.7b, qwen3:0.6b, llama3.2:1b, gemma3:1b.}

\end{figure}

Fig.~\ref{FIG:lm_tokens_per_second} shows each LM's tokens per second.
Generally, LMs with more parameters computed slower and more consistently than LMs with fewer parameters.

\begin{figure}
\begin{tabular}{c} %
	Predicted Date $\in$ Document \\ 
	\includegraphics[trim={0 0 0 0},clip,width=0.99\linewidth]{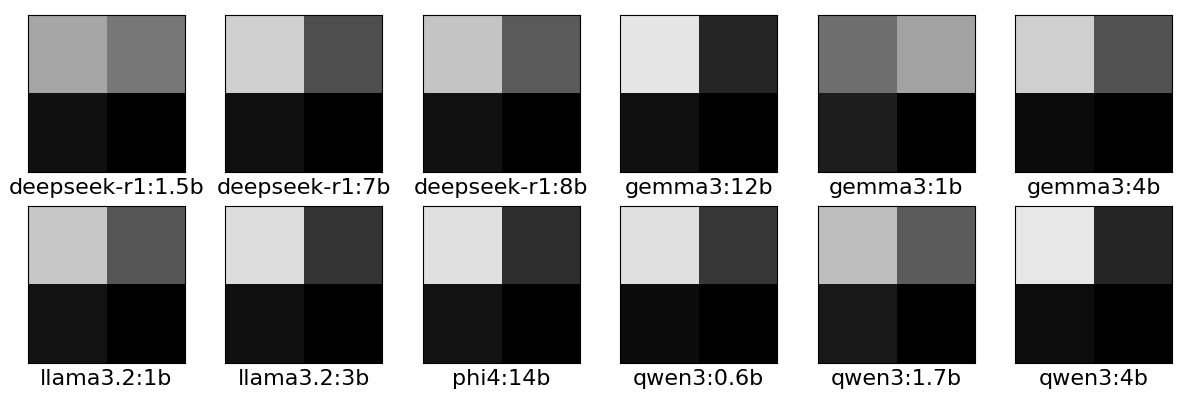} \\ %
	Predicted Date = Target \\
	\includegraphics[trim={0 0 0 0},clip,width=0.99\linewidth]{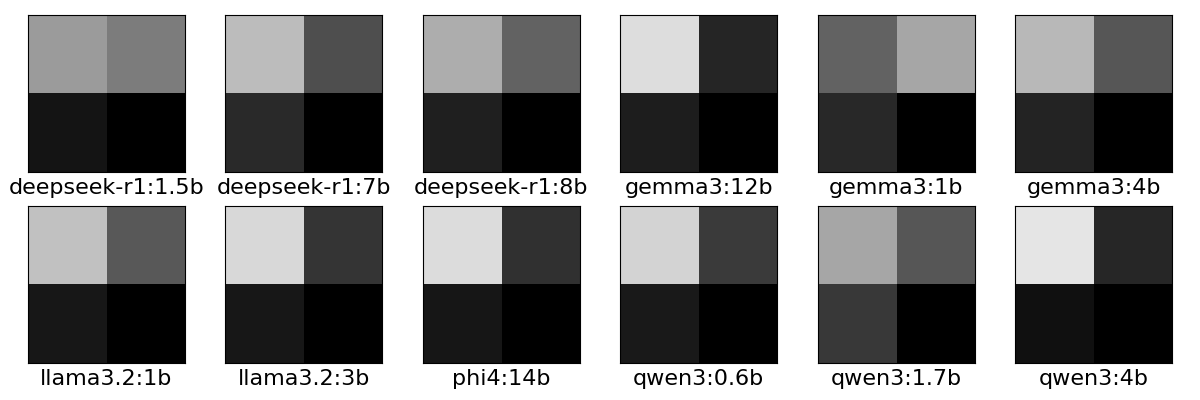} \\ 
\end{tabular}

	\caption{The predictive performance of each LM.
	The top group of confusion matrices show whether a predicted DOB exists in a document's text, and the bottom shows whether these match the respective target.
	The top-left cell in each confusion matrix shows the true positives, the top-right cell shows the false positives, the bottom-left cell shows the false negatives, and the bottom-right cell shows the true negatives (which were always 0). Lighter cells represent higher values than darker cells. 
  }
  
	\label{FIG:lm_confusion_matrices} %
	
	\Description[The predictive performance of the language models]{This figure contains (from top to bottom) a title that says ``predicted date is in document'', two rows each containing six confusion matrices, a title that says ``predicted date matches target'', two rows each containing six confusion matrices. The third and fourth rows of the confusion matrices have the same x-axis labels as the first and second rows of the confusion matrices. The first row (in left to right order) contains the labels: deepseek-r1:1.5b, deepseek-r1:7b, deepseek-r1:8b, gemma3:12b, gemma3:1b, gemma3:4b. The second row (in left to right order) contains the labels: llama3.2:1b, llama3.2:3b, phi4:14b, qwen3:0.6b, qwen3:1.7b, qwen3:4b. This figure is in greyscale. The rest of the figure is described in the caption and in Table 2.}	
\end{figure}

Fig.~\ref{FIG:lm_confusion_matrices} shows the predictive performance of each LM when these were prompted about each medical document, and another perspective of these is presented in Tab.~\ref{TAB:summarised_lm_extraction}. 
The LMs produced similar precisions to each other, and these were noticeably higher than the respective recalls. 
The range and distribution of the recalls varied significantly more than the precisions, with smaller LMs generally producing lower recalls (i.e., higher hallucination rates) than larger LMs. 
The LMs predicted a date that existed anywhere in each document more often than the respective target DOB too. 

\begin{table}

  \caption{
  Summarised knowledge extraction capability of each LM as a percentage, to 1 d.p. `-r1' was removed from the end of the deepseek-r1 models to reduce table space, and `1.5' was transformed into `$\frac{3}{2}$'. %
The first row in the table contains the major columns. The first major column (left-right) shows the language models, the second major column shows whether the predicted date was in the document, and the third major column shows whether the predicted date matched the expert-assigned target. The corresponding Precision (P), Recall (R), and F$_1$ score (F) results are shown in the 2$^{nd}$ and 3$^{rd}$ major column.
  }

\begin{tabular}{c|ccc|ccc}
  & \multicolumn{3}{c|}{prediction $\in$ document} & \multicolumn{3}{c}{prediction = target} \\ 
  \hline
  & P & R & F & P & R & F \\ 
  \hline
deepseek:$\frac{3}{2}$b & 90.8\% & 58.6\% & 71.2\% & 87.6\% & 55.8\% & 68.2\% \\
deepseek:7b & 93.2\% & 74.7\% & 82.9\% & 81.8\% & 72.1\% & 76.7\% \\
deepseek:8b & 91.6\% & 70.0\% & 79.3\% & 84.2\% & 64.5\% & 73.1\% \\
gemma3:12b & 93.4\% & 86.6\% & 89.9\% & 88.6\% & 86.0\% & 87.3\% \\
gemma3:1b & 77.9\% & 39.7\% & 52.6\% & 70.2\% & 36.4\% & 47.9\% \\
gemma3:4b & 94.7\% & 73.7\% & 82.9\% & 83.4\% & 69.5\% & 75.8\% \\
llama3.2:1b & 91.2\% & 71.6\% & 80.2\% & 88.9\% & 69.9\% & 78.3\% \\
llama3.2:3b & 93.0\% & 81.9\% & 87.1\% & 90.4\% & 81.4\% & 85.6\% \\
phi4:14b & 92.6\% & 84.1\% & 88.2\% & 91.0\% & 82.9\% & 86.8\% \\
qwen3:0.6b & 95.3\% & 81.9\% & 88.1\% & 89.2\% & 79.5\% & 84.1\% \\
qwen3:1.7b & 88.4\% & 68.8\% & 77.4\% & 75.3\% & 67.4\% & 71.1\% \\
qwen3:4b & 94.7\% & 86.8\% & 90.6\% & 93.5\% & 86.2\% & 89.7\% \\
\end{tabular}  
	\label{TAB:summarised_lm_extraction} %
\end{table}

%% file: sections/experiment_2.tex
\begin{figure}
  \centering
  Predictive Performance \& Tokens Per Second
  \\
  \includegraphics[trim={0 0 0 0},clip,width=1.0\linewidth]{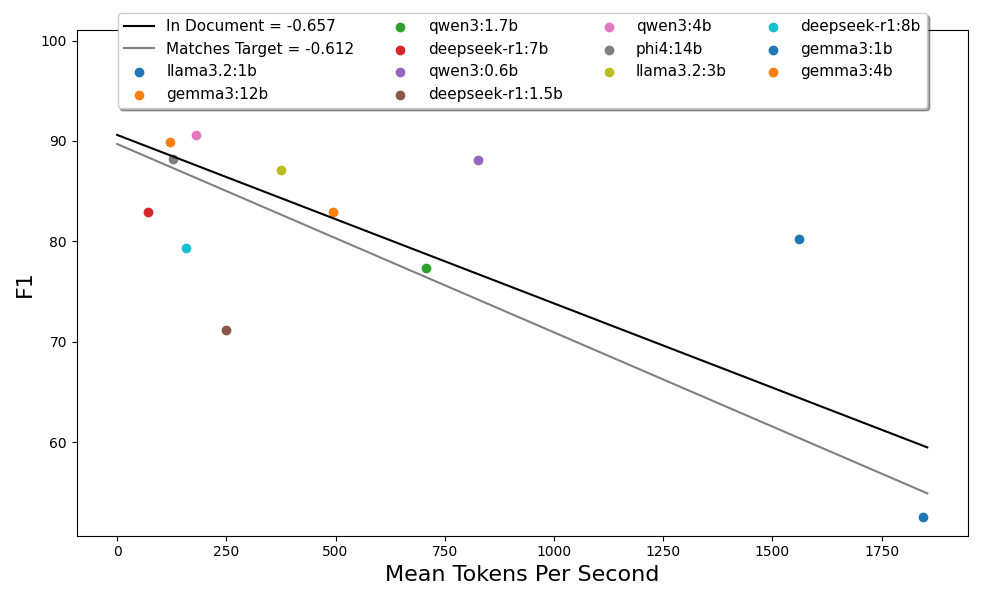}
  
  \caption{
  A contrast between each LM's F1 score and its mean tokens per second. The corresponding task of the points that were plotted was to predict whether a DOB existed in a given document. The black line indicates negative (Pearson) correlation between the mean tokens per second and whether a predicted date is contained in a document, and the grey line represents the correlation between the mean tokens per second and whether the predicted date matches the target. These lines began at the $(x,$ $y)$ point $(0,$ $max_value)$, and the slopes of these are shown in the legend.  Some of the point colours are duplicated, and other figures in this section may be required to distinguish these.
  }
  
  \label{FIG:performance_graph}
  
  \Description[A predictive performance graph.]{This is a scatterplot showing each language models F1 score and its mean tokens per second. The x-axis ranges between 0 and (approximately) 1800.  The y-axis ranges between 100 and (approximately) 50. A straight black line shows the Pearson correlation between the 12 language models, when predictive performance was measured according to whether a predicted date was in a document. The starting coordinate of this is the maximum respective F1 score and 0 tokens per second, and the end coordinate is the maximum respective F1 score multiplied by the line slope, and the maximum tokens per second. The slope of this is -0.657. A straight black line shows the Pearson correlation between the 12 language models, when predictive performance was measured according to whether a predicted date matches a document target. The starting coordinate of this is the maximum respective F1 score and 0 tokens per second, and the end coordinate is the maximum respective F1 score multiplied by the line slope, and the maximum tokens per second. This slope of this is -0.612. The language models are shown using 12 circular points. In order of furthest (visual approximation) positive distance from the black line, these are: llama3.2:1b, qwen3:0.6b, qwen3:4b, llama3,2:3b, gemma3:12b, gemma3:4b, phi4:14b, qwen3:1.7b, deepseek-r1:7b, llama3.2:1b, deepseek-r1:8b, deepseek-r1:1.5b.  
  }
\end{figure}

The second experiment applied the LMC algorithm. We began by plotting the predictive performance and tokens per second of each LM, and this is shown in Fig.~\ref{FIG:performance_graph}. This highlights the existence of strong negative correlation between LM predictive performance and computational speed. Slower LMs containing many parameters generally perform better than faster LMs that contain few. %

We then used this to produce 4 LMCs. The first and second LMs in the first LMC (chain\_1) were below the correlation line, and these (in order) were `qwen3:1.7b' and `deepseek-r1:1.5b'. The third LM in the first LMC was `qwen3:4b', and this was shown to perform quite well. The second LMC (chain\_2) was composed of 3 LMs that were furthest above the correlation line, and these (in order) were `llama3.2:1b', `qwen3:0.6b', and `qwen3:4b'. The third LMC (chain\_3) was composed of the first two LMs in the second LMC. These (in order) were `llama3.2:1b, and `qwen3:0.6b'. Finally, the fourth LMC (chain\_4) was composed (in order) of the second and first LMs in LMC 3, and these were `qwen3:0.6b' and llama3.2:1b'. Each LM in each LMC is summarised in Tab.~\ref{TAB:language_model_chain_compsition}.

\begin{table}
 \caption{The LMs that make up the 4 LMCs that were explored in this research.}

 \begin{tabular}{|c|c|} 
  \hline
  \multicolumn{2}{|c|}{Language Model Chain Composition} \\
  \hline
  chain\_1 & qwen3:1.7b, deepseek-r1:1.5b, qwen3:4b \\
  chain\_2 & llama3.2:1b, qwen3:0.6b, qwen3:4b \\
  chain\_3 & llama3.2:1b, qwen3:0.6b \\
  chain\_4 & qwen3:0.6b, llama3.2:1b \\
  \hline
 \end{tabular}
 
 \label{TAB:language_model_chain_compsition}
\end{table}

\begin{figure}
	\begin{tabular}{c}
		Predicted Date $\in$ Document\\ 
		\includegraphics[trim={0 0 0 0},clip,width=0.9\linewidth]{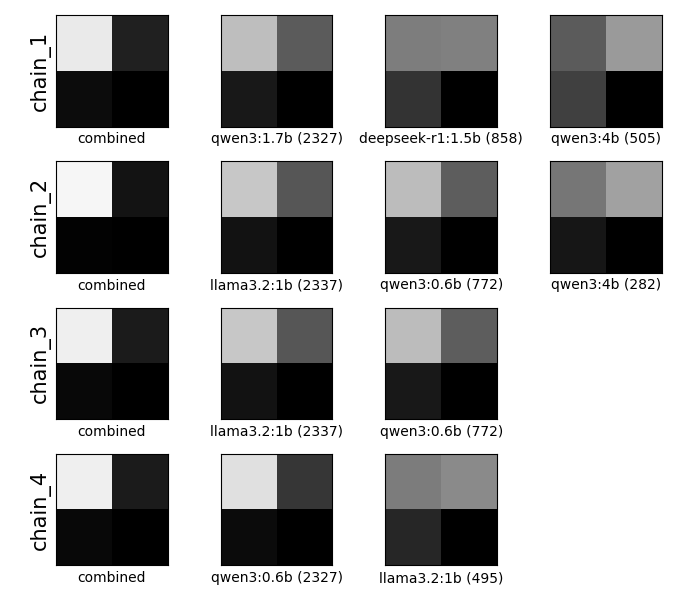} \\
		Predicted Date = Target \\  
		\includegraphics[trim={0 0 0 0},clip,width=0.9\linewidth]{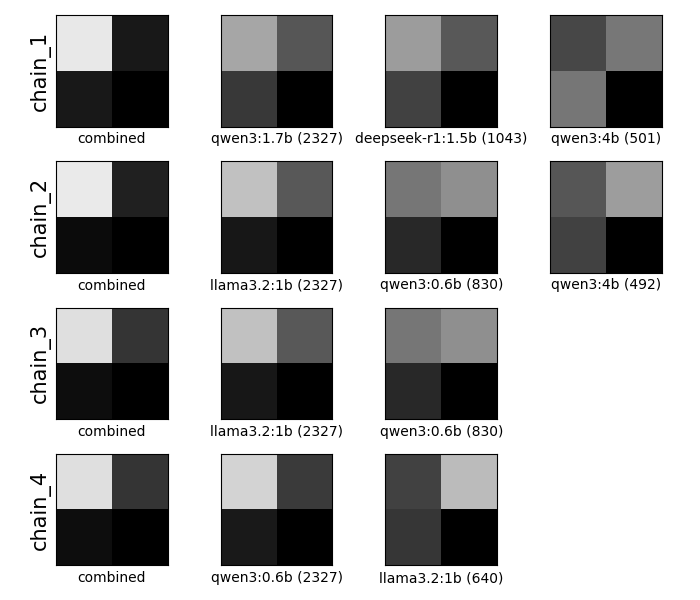} \\
	\end{tabular}

	\caption{Confusion matrices showing the predictive performance of each LMC and its component LMs. The y-axis shows the LMC that was explored, and the x-axis shows the LM.	The number in parenthesis indicates how many date examples were examined by the respective LM. `combined' indicates the cumulative predictive performance. The structure of the confusion matrices is the same as that described in Fig.~\ref{FIG:lm_confusion_matrices}. 
		}
  
	\label{FIG:chain_confusion_matrices} %
		\Description[The predictive performance of the language model chains]{
	This figure contains (from top to bottom) a title that says ``predicted date is in document'', two rows each containing four confusion matrices, two rows each containing three confusion matrices,  a title that says ``predicted date matches target'', two rows each containing four confusion matrices, and two rows each containing three confusion matrices. The first and fifth row of confusion matrices has a shared y-axis label that says chain\_1, and the x-axis labels (from left to right) are: combined, qwen3:1.7b, deepseek-r1:1.5b, and qwen3:4b. Each of these x-axis labels contains a number in parenthesis, and these are 2327, 858, and 505, for the first row, and 2327, 1043, and 501 for the fifth row. The second and sixth row of confusion matrices has a shared y-axis label that says chain\_2, and the x-axis labels (from left to right) are: combined, llama3.2:1b, qwen3:0.6b, and qwen3:4b. Each of these x-axis labels contains a number in parenthesis, and these are 2327, 772, and 282, for the second row, and 2327, 830, and 492 for the sixth row. The third and seventh rows of the confusion matrices have a shared y-axis label that says chain\_3, and the x-axis labels (from left to right) are: combined, llama3.2:1b, and qwen3:0.6b. Each of these x-axis labels contains a number in parentheses, and these are 2327 and 772, for the third row, and 2327 and 830 for the seventh row. The fourth and eighth row of confusion matrices has a shared y-axis label that says chain\_4, and the x-axis labels (from left to right) are: combined, qwen3:0.6b, and llama3.2:1b. Each of these x-axis labels contains a number in parenthesis, and these are 2327, 772, and 282, for the fourth row, and 2327 and 640 for the eighth row. This figure is in greyscale. The rest of the figure is described in the caption and in Table 4.}	
\end{figure}

Fig.~\ref{FIG:chain_confusion_matrices} shows the predictive performance of each LMC and each of its respective LMs when these were prompted about the DOB in each medical document using the same prompt from before, and a summarised perspective of these results is shown in Tab.~\ref{TAB:summarised_chain_extraction}. 
The combined predictive performance of each LMC greatly exceeded the performance of its individual LM components, and these are much higher than the predictive performance of the LMs that were described in Tab.~\ref{TAB:summarised_lm_extraction}. 
The LMCs predicted more dates that existed anywhere in each document than they predicted target dates too.
The results reveal that many of the DOBs were successfully identified by the first LM in each chain, and that later LMs in an LMC were more likely to hallucinate than LMs which occurred earlier. 
Finally, LMCs 3 and 4 produced the same predictive results, which suggests that the predictive performance of LMs in an LMC is commutative. 

\begin{table}

  \caption{
  Summarised knowledge extraction capability of each LMC and its component LMs as a percentage, to 1 d.p. `deepseek-r1:1.5b' has been transformed to `deepseek:$\frac{3}{2}$' to save space in the table. This is a predictive performance summary of the results in Fig.~\ref{FIG:chain_confusion_matrices}. The meanings of the columns in this table are the same as described in Tab.~\ref{TAB:summarised_lm_extraction}, except that the first major column indicates LMCs and LMs. The LMCs (chain\_1-4) are written in bold.
    }

\begin{tabular}{c|ccc|ccc}
  & \multicolumn{3}{c|}{prediction $\in$ document} & \multicolumn{3}{c}{prediction = target} \\ 
  \hline
  & P & R & F & P & R & F \\ 
  \hline
\textbf{chain\_1} & 95.1\% & 88.4\% & 91.6\% & 91.0\% & 90.8\% & 90.8\% \\
   \hline
qwen3:1.7b & 88.4\% & 68.8\% & 77.4\% & 75.3\% & 67.4\% & 71.1\% \\
deepseek:$\frac{3}{2}$b & 71.5\% & 49.2\% & 58.3\% & 71.6\% & 65.5\% & 68.4\% \\
qwen3:4b & 58.9\% & 35.9\% & 44.6\% & 36.1\% & 35.9\% & 36.0\% \\
    \hline
\textbf{chain\_2} & 99.0\% & 93.2\% & 96.0\% & 95.2\% & 88.5\% & 91.8\% \\
   \hline
 llama3.2:1b & 91.2\% & 71.6\% & 80.2\% & 88.9\% & 69.9\% & 78.3\% \\
 qwen3:0.6b & 88.1\% & 67.9\% & 76.7\% & 74.2\% & 44.8\% & 55.9\% \\
 qwen3:4b & 84.0\% & 42.1\% & 56.1\% & 56.7\% & 33.8\% & 42.3\% \\
   \hline
\textbf{chain\_3} & 96.9\% & 90.0\% & 93.3\% & 94.2\% & 82.1\% & 87.7\% \\
  \hline
 llama3.2:1b & 91.2\% & 71.6\% & 80.2\% & 88.9\% & 69.9\% & 78.3\% \\
 qwen3:0.6b & 88.1\% & 67.9\% & 76.7\% & 74.2\% & 44.8\% & 55.9\% \\
   \hline
\textbf{chain\_4} & 96.9\% & 90.0\% & 93.3\% & 94.2\% & 82.1\% & 87.7\% \\
  \hline
 qwen3:0.6b & 95.3\% & 81.9\% & 88.1\% & 89.2\% & 79.5\% & 84.1\% \\
 llama3.2:1b & 75.7\% & 47.2\% & 58.2\% & 54.1\% & 25.0\% & 34.2\% \\

\end{tabular}
  
	\label{TAB:summarised_chain_extraction} %
\end{table}

\begin{figure}
  \centering
  LMC Tokens Per Second
  \\
  \includegraphics[trim={0 0 0 0},clip,width=1.0\linewidth]{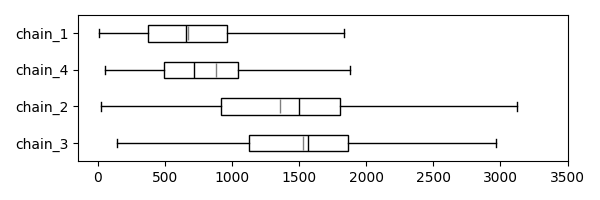}
  
  \caption{
	The tokens per second resulting from each LMC. The figure properties are the same as in Fig.~\ref{FIG:lm_tokens_per_second}, except that the y-axis labels (from top to bottom) are: chain\_1, chain\_4, chain\_2, and chain\_3 
  }
  
  \label{FIG:chain_tokens_per_second}
  
  \Description[The language models tokens per second]{Figure fully described in the caption.}
\end{figure}

\begin{figure}
  \centering
  Reciprocal of Mean Tokens Per Second 
  \includegraphics[trim={0 0 0 0},clip,width=1.0\linewidth]{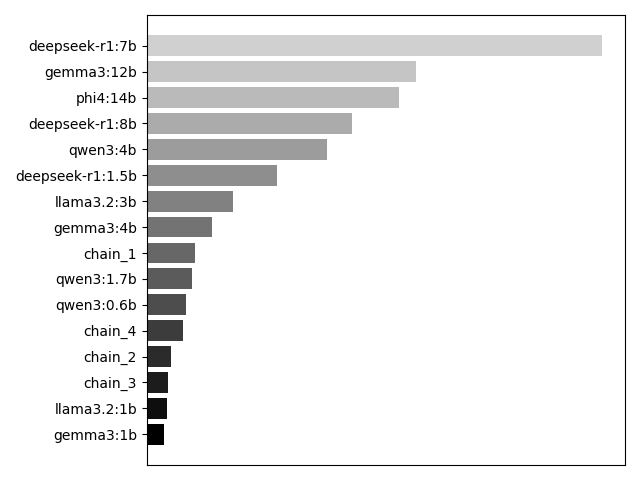}
  
  \caption{
  The reciprocal of each LM and LMC's mean tokens per second. This is an estimate of how long each of these would take to extract the DOB from an arbitrary number of medical documents. The time required increases from left to right, and the execution speed increases in descending order. Lighter bars represent higher values than darker bars. 
  }
  
  \label{FIG:multiplicative_inverse_of_tokens_per_second}
  
  \Description[The reciprocal of each language model and language model chain's mean tokens per second]{
  The majority of the figure is described in its caption.
  However the y-axis labels (from top to bottom) are: deepseek-r1:7b, gemma3:12b, phi4:14b, deepseek-r1:8b, qwen3:4b, deepseek-r1:1.5b, llama3.2:3b, gemma3:4b, chain\_1, qwen3:1.7b, qwen3:0.6b, chain\_4, chain\_2, chain\_3, llama3.2:1b, gemma3:1b.}
  
\end{figure}

Fig.~\ref{FIG:chain_tokens_per_second} shows the token per second results of each LMC, and Fig.~\ref{FIG:multiplicative_inverse_of_tokens_per_second} shows the reciprocal of each LM's and LMC's mean tokens per second.
The range of each LMC's tokens per second result was visibly larger than the corresponding interquartile range.
LMCs are often significantly faster than the individual LMs, and the order of the LMs in these appears to be significant.
For instance, chain\_3 was much faster than chain\_4, despite containing the same LMs and producing the same predictions.

%% file: sections/discussion.tex
\subsection{Language Model Chains Are Faster \& More Predictive Than Their Components}
Our results show that the LMCs were able to extract the DOBs from medical documents more often and much faster than the individual LMs that they were composed of, and LMCs appear to greatly reduce generative hallucinations. 
Our initial hypothesis was that LMCs would be faster, but a large boost to predictive performance was unexpected. 
This is reminiscent of boosted ensemble learning \cite{breiman2001random}, where the combined performance of several weak learners exceeds that of a strong learner, and a worthwhile endeavour would be to explore how techniques and properties in this domain could be applied to the LMC algorithm in the future.

\subsection{The Order Of Language Models In A Chain Matters}
LMC 3 and LMC 4 suggest that the order of LMs in an LMC may significantly impact computational speed, but not necessarily predictive performance. 
It was not possible to exhaustively explore the many possible LM combinations and permutations in this research, but we highly recommend further exploration of this in the future.

\subsection{More Parameters Do Not Guarantee Better Predictive Accuracy}
The largest and slowest LMs did not necessarily extract DOBs more precisely than smaller and faster LMs, although these are often publicised as performing better on benchmark datasets \cite{wang2025enterprise}. 
This suggests that there is a lot of variability in how an off-the-shelf LM will perform on a given task. %

%% file: sections/conclusion.tex
Our task was to extract Dates Of Births (DOBs) from medical documents that  contained several competing answers. 
We were required to use a local device that used CPUs instead of GPUs, and the documents contained sensitive data that needed to be handled carefully and securely.
We achieved our task by proposing, implementing, and applying the novel Language Model Chain (LMC) algorithm.
This conceptually-simple algorithm used multi-stage cascade predictions, candidate answers, and sequential application of `\textit{good enough}' Language Models (LMs) to extract Dates Of Birth (DOB) from text that contained multiple other dates. 
It was rarely possible to extract the target DOB without contextual information, and our results found that LMCs exceeded LMs in both predictive performance and reduced computational cost during this task.
We believe that this research contributes greatly to the problems of knowledge extraction and information retrieval.

We highlighted opportunities for further exploration and development in the future. Namely: 
more exploration and exposure of the LMC algorithm and its component LMs,
whether the LMC algorithm performs on hardware other than a CPU,
how the ambiguous prompt explored in this research impacted the findings,
whether it is possible to extract candidate answers from data that does not contain an explicit pattern,
and better methods for identifying the optimal LMC structure. %

%% file: sections/acknowledgements.tex
We would like to thank all of our colleagues from TMLEP, Elysium Web Services, and the University of Kent who were involved in this research.
Particularly to Rebecca Dole, Steve Hawes, and Wendy Hawes for proofreading this research paper.
This research was partially funded by a UK Research and Innovation (UKRI) grant (grant number 10048265) that was awarded to a knowledge transfer partnership project, whose goal was to modernise medical data pagination.